%% file: main.tex
\documentclass[10pt,twocolumn,letterpaper]{article}

\usepackage{wacv}      


\definecolor{wacvblue}{rgb}{0.21,0.49,0.74}
\usepackage[pagebackref,breaklinks,colorlinks,allcolors=wacvblue]{hyperref}

\def\wacvPaperID{585} 
\def\confName{wacv}
\def\confYear{2026}

\title{Multimodal Semantic-Aware Contrastive Learning For False Negative Mitigation in 3D Medical Imaging}

\author{Sara Ketabi\\ University of Toronto\\ The Hospital for Sick Children\\Vector Institute
\and
Matthias W. Wagner\\The Hospital for Sick Children\\University Hospital Augsburg
\and
Cynthia Hawkins\\The Hospital for Sick Children
\and
Uri Tabori\\The Hospital for Sick Children
\and
Birgit Betina Ertl-Wagner\\University of Toronto\\The Hospital for Sick Children
\and
Farzad Khalvati\\University of Toronto\\The Hospital for Sick Children\\Vector Institute}

\begin{document}
\maketitle
\input{sec/0_abstract}    
\input{sec/1_intro}

\input{sec/2_formatting}
\input{sec/3_finalcopy}

\input{sec/4_results}

\input{sec/5_discussion}

\input{sec/6_conclusion}
{
    \small
    \bibliographystyle{ieeenat_fullname}
    \bibliography{main}
}

\end{document}

%% file: sec/0_abstract.tex
\begin{abstract}
Multimodal Contrastive Learning (CL) has shown significant performance in aligning representations across various data modalities and improving  downstream tasks, especially in healthcare. It works by minimizing the distance between matched (positive) data modalities, while maximizing the distance between mismatched (negative) samples. Traditional CL frameworks typically assume instance-based correspondence within data batches, treating all non-paired samples as negatives. However, this assumption often fails in medical settings, where samples may share high-level semantic attributes, leading to false negatives that degrade representation quality. In this paper, we propose Multimodal Semantic-Aware Contrastive Learning (MseaCL), a CL framework trained on a pediatric cohort of 3D brain magnetic resonance imaging (MRI) scans and radiology reports. The goal of this framework is to mitigate the impact of semantically similar false negative samples by incorporating semantic similarity between radiology reports, as a guiding signal during the learning process. Our results indicate that applying this framework as a pretraining stage can achieve notable improvements in downstream tasks, e.g., at least a 22.6\% increase in the area under the receiver operating characteristic curve (AUC) of pediatric brain tumor molecular classification, demonstrating its potential for more robust and semantically aligned multimodal representations in clinical applications.
\end{abstract}

%% file: sec/1_intro.tex
\section{Introduction}
\label{sec:intro}

Contrastive learning (CL) has emerged as a powerful framework for representation learning, particularly when applied to multimodal settings. In medical imaging, multimodal CL frameworks that jointly leverage images and associated clinical text, such as radiology reports, have demonstrated substantial improvements over unimodal approaches \cite{ketabi2025multimodal}. These gains result from the ability of multimodal CL to use complementary information across modalities, enabling the model to capture both low-level visual patterns and high-level semantic concepts.

In a typical multimodal CL framework, the learning objective encourages the representations of corresponding image–report pairs (positive pairs) to be close in the embedding space, while simultaneously pushing apart representations of non-corresponding image–report pairs (negative pairs) sampled from the same training batch. Through this alignment mechanism, the model learns to associate visual features in medical images with clinically meaningful semantic concepts described in reports. As a result, the learned image representations tend to be more informative, transferable, and better suited for downstream clinical tasks.

However, this formulation relies on an implicit assumption that all mismatched image–report pairs constitute true negatives, an assumption that is often not true in real-world medical datasets. In practice, a training batch may include image–report pairs from different patients that share significant semantic similarity, such as comparable pathological findings, anatomical regions, or disease subtypes. Treating such semantically related samples as negatives introduces false negatives into the contrastive objective, forcing the model to separate representations that should remain close in the latent space.

The presence of false negatives can have several negative effects. First, it may degrade the semantic structure of the learned representation space by ignoring clinically meaningful similarities across patients. Second, it can negatively impact downstream task performance, particularly for tasks that rely on subtle semantic distinctions. Finally, especially in the medical domain, false negatives may undermine model explainability, as the learned representations may not emphasize clinically relevant image regions or features aligned with expert knowledge.

To mitigate these issues, prior studies have proposed various strategies for identifying and handling false negatives in CL frameworks. These include clustering-based approaches in the embedding space, similarity-based heuristics, and adaptive reweighting or removal of suspected false negatives from the contrastive loss \cite{huynh2022boosting,chen2021incremental,xu2024contrastive}. While promising, most of these methods have been developed for text-only domains or 2D imaging scenarios, where the data structure is relatively simple. Extending such approaches to 3D medical imaging, such as brain MRI, remains challenging due to several reasons, including higher dimensionality and complex spatial structure.

In this work, we introduce a semantic-aware multimodal CL framework for 3D brain MRI scans and their associated radiology reports. The key idea is to incorporate report-level semantic similarity directly into the CL objective. We compute the cosine similarity between report embeddings and use this similarity as a semantic reference to determine the contribution of mismatched image–report pairs for a given anchor sample. To that end, the framework reduces the penalization of semantically similar pairs that would otherwise be treated as hard negatives, allowing shared clinical information to be preserved in the learned representations.

By explicitly accounting for semantic relationships between reports, the proposed framework encourages the model to encode clinically meaningful similarities in the image embedding space. The resulting image representations would be more semantically coherent and transferable, leading to improved performance on downstream tasks such as brain tumor genetic marker classification. Moreover, these representations enhance model explainability by encouraging reliance on task-relevant image features, thereby increasing alignment between model decision-making and clinical reasoning.

The main contributions of this chapter are summarized as follows:

\begin{itemize}
    \item We identify and analyze the impact of false negatives in sample-based multimodal CL for 3D medical imaging, highlighting their negative effects on both downstream task and explainability.
    \item We propose a semantic-aware CL framework that leverages report-level semantic similarity to adaptively weight mismatched image–report pairs during training.
    \item We demonstrate the effectiveness of the proposed framework on downstream 3D brain MRI tasks, showing consistent improvements in model performance and explainability compared to conventional multimodal baselines.
    
\end{itemize}

%% file: sec/2_formatting.tex
\section{Related Work}
\label{sec:formatting}

Typical multimodal instance-based CL frameworks, such as Contrastive Language-Image Pre-training (CLIP) \cite{radford2021learning}, rely on one-to-one correspondences between images and associated reports. In the medical domain, similar architectures have been applied to image–report CL, primarily in 2D modalities such as chest X-rays \cite{huang2021gloria,wang2022multi,ji2021improving}. These approaches generally treat all non-matching pairs as negatives regardless of their latent semantic similarity, which introduces the false negative problem, particularly in domains with  overlapping semantics such as medical imaging.

A number of studies have proposed strategies to detect or mitigate false negatives in CL. Huynh et al. \cite{huynh2022boosting} introduced a support-set–based method that identifies false negatives by aggregating similarity scores between a set of support views and negative samples. Negative instances with high aggregated similarity are either removed or relabeled as positives, yielding improved performance on several ImageNet\cite{deng2009imagenet}-based benchmarks. Xu et al. \cite{xu2024contrastive} addressed false negatives in text contrastive learning by computing cosine similarity between anchor and augmented negative sentence representations. Negative samples exceeding a similarity threshold were downweighted through adaptive weighting in the loss function, leading to improvements on multiple semantic textual similarity datasets.

Clustering-based approaches have also been explored. Chen et al. \cite{chen2021incremental} proposed detecting false negatives by performing k-means clustering on image features to obtain pseudo labels. Images assigned to the same cluster centroid were treated as semantically related and thus removed from the negative group or converted to positives, improving Top-1 accuracy on datasets such as ImageNet. Other work has incorporated class-level or semantic structure to reduce the likelihood of false negatives, including supervised-style contrastive learning \cite{robinson2020contrastive}, where class labels were used to guide the selection of negative samples.


Within the medical domain, several vision–language CL frameworks have integrated semantic information to address false negatives or noisy correspondences. Wang et al. \cite{wang2022medclip} proposed MedCLIP, which replaces the traditional InfoNCE loss with a semantic matching objective based on soft similarity targets between unpaired chest X-ray images and reports. This approach achieves improved performance over CLIP and GLORIA \cite{huang2021gloria}, a multimodal CL framework for that aligns global and local representations of chest X-ray images and radiology reports, across a range of medical imaging tasks.

In spite of effectiveness, existing methods for mitigating false negatives predominantly focus on textual or 2D visual data. To the best of our knowledge, no prior work has addressed false-negative mitigation in multimodal CL involving 3D medical imaging. This gap is particularly important because 3D images contain richer spatial information, making  negative sampling more critical to accurately encode shared semantic information among samples. 

%% file: sec/3_finalcopy.tex
\section{Method}

\begin{figure*}[t]
  \centering 
      \includegraphics[width=4in]   
      {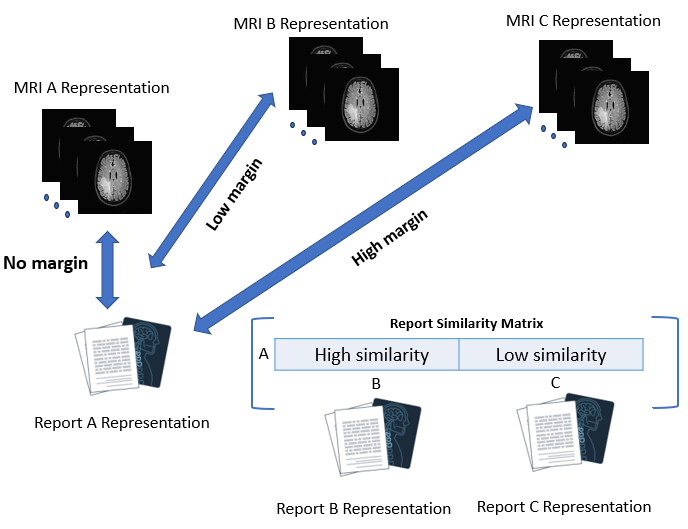} 
      \caption{Overview of False Negative Mitigation in MseaCL. For each mismatched MRI-report pair, the semantic similarity between corresponding radiology reports adaptively determines the margin used in the contrastive loss, where a higher margin encourages larger separation.} 
      \label{seacl} 
    \end{figure*}

In this work, we propose Multimodal Semantic-aware Contrastive Learning (MSeaCL), a contrastive framework designed to detect and mitigate the effect of false negatives in multimodal CL using paired 3D MRI scans and radiology reports. The underlying assumption is that two MRI-report pairs can convey highly similar clinical findings despite originating from different patients. Treating such semantically similar samples as strong negatives pushes them apart in the latent space, undermining both representation learning and downstream task performance. To address this limitation, MSeaCL assigns non-uniform weights to negative samples based on their semantic similarity. Intuitively, negative pairs that exhibit high semantic similarity should be penalized less strongly, while semantically unrelated samples should be subjected to the full contrastive penalty. Figure \ref{seacl} illustrates the architecture of the proposed framework.

Our proposed framework operates by aligning two levels of embeddings, global and local, extracted from both image and text encoders, as depicted in Figure in the supplementary material. The global embeddings capture high-level, aggregated information for each modality. Specifically, the global MRI embedding was obtained from the final layer of the MRI encoder, while the global report embedding was extracted from the last layer of the text encoder by taking the embedding associated with the [CLS] token.

The local embeddings, in contrast, preserve fine-grained and spatially or contextually detailed information. The local MRI embedding was taken from the second-to-last layer of the image encoder, whereas the local report embedding was derived from the last transformer layer but excludes the [CLS] token, retaining only the embeddings of the remaining tokens. We applied a cross-attention module to generate weighted local representations for each modality. Specifically, to obtain the weighted local text representation, the local text features were used as the query, while the key and value inputs were derived from the local MRI representation; the same process was performed in reverse to obtain the weighted local MRI representation. The local contrastive loss was then computed using both the original local representations and their corresponding cross-attention–weighted counterparts.

Given a batch of $N$ image-text pairs, for global-level representation alignment, let $\mathbf{I} \in \mathbb{R}^{N \times d}$ denote global image embeddings and $\mathbf{T} \in \mathbb{R}^{N \times d}$ denote global text embeddings, where $d$ is the embedding dimensionality. 
The embeddings were L2-normalized before computing similarities to ensure numerical stability. The normalized embeddings were then used in all subsequent computations.

\begin{equation}\hat{\mathbf{I}}_i = \frac{\mathbf{I}_i}{\|\mathbf{I}_i\|_2}, \quad \hat{\mathbf{T}}_i = \frac{\mathbf{T}_i}{\|\mathbf{T}_i\|_2}
\end{equation}

To capture semantic relationships between samples, we computed pairwise cosine similarity between corresponding report embeddings. The cosine similarity between two embeddings $\mathbf{T}_i$ and $\mathbf{T}_j$ can be represented as:

\begin{equation}
\mathbf{S}_{ij}=\text{cos}(\mathbf{T}_i, \mathbf{T}_j) = \frac{\mathbf{T}_i \cdot \mathbf{T}_j}{\|\mathbf{T}_i\|_2 \|\mathbf{T}_j\|_2} = \frac{\sum_{k=1}^{d} T_{ik} T_{jk}}{\sqrt{\sum_{k=1}^{d} T_{ik}^2} \sqrt{\sum_{k=1}^{d} T_{jk}^2}}
\end{equation}



To focus on negative pairs (non-matching pairs), we mask the diagonal elements (positive pairs) by setting them to $-\infty$:

\begin{equation}
\tilde{\mathbf{S}}_{ij} = \begin{cases}
-\infty & \text{if } i = j \\
\mathbf{S}_{ij} & \text{otherwise}
\end{cases}\end{equation}

The masked semantic similarity was then normalized using softmax to obtain a probability distribution over negative pairs:

\begin{equation}\bar{\mathbf{S}}_{ij} = \frac{\exp(\tilde{\mathbf{S}}_{ij})}{\sum_{k=1}^{N} \exp(\tilde{\mathbf{S}}_{ik})}\end{equation}


The contrastive logits between image and text embeddings, which capture the alignment between them, were computed as the dot product of normalized embeddings:

\begin{equation}\mathbf{M}_{ij} = \mathbf{I}_i^T \mathbf{T}_j\end{equation}


Traditional margin-based contrastive losses \cite{schroff2015facenet} use a fixed margin for all negative pairs. However, semantically similar negative pairs should be treated differently from actual dissimilar pairs. Here, we introduce an adaptive margin that is adjusted based on semantic similarity:

\begin{equation}m_{ij} = m_{\text{base}} \cdot (1 - \bar{\mathbf{S}}_{ij})\end{equation}

where $m_{\text{base}}$ is the base margin parameter. This formulation ensures that: When $\bar{\mathbf{S}}_{ij} \approx 1$ (high semantic similarity), $m_{ij} \approx 0$, allowing semantically similar negatives to be closer in the embedding space, and when $\bar{\mathbf{S}}_{ij} \approx 0$ (low semantic similarity), $m_{ij} \approx m_{\text{base}}$, enforcing the standard margin for actual dissimilar pairs.

Our loss function consists of two components: positive pair loss and negative pair loss, computed bidirectionally for both image-to-text and text-to-image directions. For positive pairs (diagonal elements), the logits were encouraged to be close to 1:

\begin{equation}\mathcal{L}_{\text{pos}}^{\text{i2t}} = \sum_{i=1}^{N} (1 - \mathbf{M}_{ii})^2\end{equation}

\begin{equation}\mathcal{L}_{\text{pos}}^{\text{t2i}} = \sum_{i=1}^{N} (1 - \mathbf{M}_{ii})^2\end{equation}

For negative pairs, the logits were encouraged to be below the adaptive margin. The loss was then computed using $\text{ReLU}(x) = \max(0, x)$ to ensure that only logits exceeding the adaptive margin contribute to the loss.

\begin{equation}\mathcal{L}_{\text{neg}}^{\text{i2t}} = \sum_{i=1}^{N} \sum_{j \neq i} \left[\text{ReLU}(\mathbf{M}_{ij} - m_{ij})\right]^2\end{equation}

\begin{equation} 
\mathcal{L}_{\text{neg}}^{\text{t2i}} = \sum_{i=1}^{N} \sum_{j \neq i} \left[\text{ReLU}(\mathbf{M}_{ji} - m_{ij})\right]^2\end{equation}


The total loss consists of positive and negative losses for both directions:

\begin{equation}\mathcal{L}_{\text{i2t}} = \frac{1}{N} \left(\mathcal{L}_{\text{pos}}^{\text{i2t}} + \mathcal{L}_{\text{neg}}^{\text{i2t}}\right)\end{equation}

\begin{equation}\mathcal{L}_{\text{t2i}} = \frac{1}{N} \left(\mathcal{L}_{\text{pos}}^{\text{t2i}} + \mathcal{L}_{\text{neg}}^{\text{t2i}}\right)\end{equation}

The global bidirectional loss was computed as the average of both directions:

\begin{equation}\mathcal{L}_{\text{global}} = \frac{1}{2} \left(\mathcal{L}_{\text{i2t}} + \mathcal{L}_{\text{t2i}}\right)\end{equation}

For local-level representation alignment, the same contrastive framework was applied at two stages: 1. Local image alignment, with $\mathbf{I}$ indicating original local image embeddings and $\mathbf{T}$ indicating  cross-attention-weighted local image embeddings. 2. Local text alignment, with $\mathbf{I}$ and $\mathbf{T}$ demonstrating original local text embeddings and cross-attention-weighted local text embeddings, respectively. At each stage, we computed the bidirectional contrastive loss using the same formulation as the global loss, guided by the semantic similarity matrix derived from global text embeddings. The local loss aggregates both stages:

\begin{equation}\mathcal{L}_{\text{local}} =  (\mathcal{L}_{\text{local-image-alignment}} + \mathcal{L}_{\text{local-text-alignment}})\end{equation}

Finally, the total loss was calculated as the weighted summation of global and local losses, where the weights ($\alpha$ and $\beta$) are trainable parameters.

\begin{equation}
     \mathcal{L}_{\text{total}} = \alpha \times\mathcal{L}_{\text{global}} + \beta \times \mathcal{L}_{\text{local}}
\end{equation}













The base margin $m_{\text{base}}$ was set to 0.5, though it can be tuned based on the specific task and dataset characteristics.

These encoders produce two levels of representations—global and local. The global representations capture high-level, aggregated information for each modality. Specifically, the global MRI representation is obtained from the final layer of the MRI encoder, while the global report representation is extracted from the last layer of the text encoder by taking the embedding associated with the [CLS] token.

The local representations, in contrast, preserve fine-grained and spatially or contextually detailed information. The local MRI representation is taken from the second-to-last layer of the image encoder, whereas the local report representation is derived from the last transformer layer but excludes the [CLS] token, retaining only the embeddings of the remaining tokens.

The similarity matrix computed from the global report representations was also used to guide the local-level contrastive loss, as it provides a reliable estimate of semantic similarity between samples. Before computing the local loss, we applied a cross-attention module to generate weighted local representations for each modality. Specifically, to obtain the weighted local text representation, the local text features were used as the query, while the key and value inputs were derived from the local MRI representation; the same process was performed in reverse to obtain the weighted local MRI representation. The local contrastive loss was then computed using both the original local representations and their corresponding cross-attention–weighted counterparts.

%% file: sec/4_results.tex
\section{Experiments and Results}

\subsection{Downstream Tasks}




To evaluate the effectiveness of MSeaCL, we fine-tuned the pretrained image encoder on a clinically important downstream task: three-class molecular diagnosis of pLGG. This task focuses on distinguishing between key molecular subtypes, i.e., BRAF Fusion, BRAF V600E Mutation, and Other. 

We evaluated the proposed framework from two complementary perspectives. First, we assessed classification performance to determine whether MSeaCL improves multi-class classification performance and generalization. Second, we analyzed visual explainability by examining the alignment between model attention and tumor regions, providing insights into whether the model bases its predictions on clinically meaningful features.

\subsection{Data}
Our internal dataset comprised 341 paired MRI (FLAIR sequence)–report samples collected from a pediatric hospital, spanning 2000 to 2018. These image–text pairs were used to pretrain MSeaCL, enabling the model to learn semantically aligned multimodal representations. For downstream fine-tuning, we further incorporated biopsy-derived genetic-marker labels as the classification ground-truth, which were evenly distributed as: BRAF Fusion: , BRAF Mutation: , and Other: . Furthermore, radiologist-annotated tumor segmentation masks were applied only to quantitatively assess explainability.

To test model generalizability, we employed an independent cohort from a different time period, i.e., between 2018 and 2023, using different MRI scanners. This external dataset included 99 MRI scans, each with corresponding genetic-marker labels and tumor segmentation masks. The dataset shift across scanners and acquisition times provided a robust setting to assess the stability of the learned representations and the performance of the downstream classifier.

\subsection{Implementation Details}


We employed a 3D ResNet architecture \cite{he2016deep} augmented with a self-attention module \cite{vaswani2017attention} as the image encoder to better capture long-range spatial dependencies within volumetric MRI scans. As the text encoder, we used a Longformer transformer \cite{Beltagy2020Longformer}, initialized with weights from Clinical Longformer \cite{li2022clinical}, enabling efficient processing of long radiology reports while leveraging domain-specific text representations.

The MSeaCL framework was pretrained for 400 epochs. For the downstream pLGG molecular classification task, we fine-tuned only the pretrained image encoder for five epochs, demonstrating the efficiency of the MseaCL pretraining. During fine-tuning, the first two convolutional layers were frozen, while the last two convolutional layers and the self-attention module were updated.

During pretraining, we used a batch size of 16, which was selected to accommodate the memory constraints imposed by 3D volumetric data processing. The base contrastive margin was set to 0.5. AdamW \cite{loshchilov2017decoupled} was chosen as the optimizer for its effective decoupling of weight decay. For the downstream classifier, we used Cross-Entropy Loss with a label-smoothing factor of 0.1 and class-proportional weights to mitigate class imbalance.

To stabilize the training across the different components of the architecture, we applied separate learning rates: 1e-4 for the convolutional backbone and 5e-4 for the classification head, to enable larger updates in randomly initialized  layers. A StepLR learning-rate scheduler was used during fine-tuning, reducing the learning rate by a factor of 0.5 every 10 epochs to encourage convergence and prevent overfitting.

\subsection{Evaluation Setting}



To comprehensively evaluate both classification performance and model explainability, we compared MSeaCL against three baselines. All baselines used the same classification architecture, i.e., a 3D ResNet with a self-attention module and two fully connected layers, to ensure a fair comparison focused on the effect of representation initialization. The baselines differed in how the image encoder was initialized: 1. Random initialization, 2. Initialization with Med3D segmentation-based pretrained weights \cite{chen2019med3d}, and 3. Initialization with a conventional instance-based CL framework \cite{ketabi2025multimodal}, proposed in Chapter 5.
This setting allowed us to compare the effect of semantic-aware pretraining in MSeaCL to widely used initialization techniques. It is worth noting that CLIP \cite{radford2021learning}, SimCLR, and GLORIA \cite{huang2021gloria} were not included as baselines, as they are specifically designed for 2D image data and are therefore not directly comparable to our 3D MRI setting.

For classification performance, we used AUC as the primary evaluation metric due to its relevance for clinical decision-making. We additionally reported precision, recall, and F1-score to capture complementary aspects of performance and provide a more complete assessment of model behavior across classes. To assess explainability, we quantified the alignment between model attention and ground-truth tumor segmentation masks. Specifically, we computed the Dice similarity coefficient between the self-attention–derived saliency maps and segmentation masks. A threshold of 0.05 was used to binarize the model attention maps. This was performed using both 3D (entire MRI volume) and 2D (slice containing the largest tumor cross-section).

For internal evaluation, we performed five-fold cross-validation, training five separate downstream models and reporting the mean test performance across folds. Since each datapoint corresponded to a unique patient ID, all data splits were performed at the patient level, ensuring that no patient appeared in more than one of the training, validation, or test sets. For external evaluation, we applied all five fine-tuned models to the independent external dataset and averaged their results to obtain the final metrics, improving the reliability of the reported findings. Furthermore, there was no overlap between the patients in the internal and external datasets, ensuring a rigorous assessment of out-of-distribution generalization.

\subsection{Classification Performance Results}


Tables \ref{table2} and \ref{table3} summarize the internal and external classification performance of the 3D ResNet model under different initialization strategies. Overall, the results demonstrate that pretraining with MSeaCL consistently leads to better downstream diagnostic performance, with particularly significant improvements in out-of-distribution settings.


In the internal evaluation, where the training and test data originate from the same imaging distribution, the model initialized with MSeaCL achieved the highest AUC among all compared approaches. Although the improvement over the strongest baseline, i.e., instance-based CL, was modest, it remains meaningful because internal performance can often reach an upper bound when models capture dataset-specific data patterns.

In contrast, the external evaluation revealed significantly larger performance differences. When tested on data acquired from a different time period and using different MRI scanners, which can cause substantial domain shifts, MSeaCL indicated remarkable performance gains. Notably, it achieved a 0.22 increase in AUC compared to the instance-based CL baseline, underscoring the benefit of incorporating semantic alignment during pretraining. Beyond AUC, MSeaCL outperformed all baselines by large margins across every reported metric, indicating superior generalization and reliability in scenarios where distribution shifts can happen.

 Overall, these findings demonstrate that semantic-aware multimodal pretraining not only provides marginal gains in internal settings but also yields significant improvements in model generalizability in external validation, which is essential for real-world adoption in clinical applications.

\begin{table*}
  \centering 
  \caption{Mean Internal Test Classification performance results of 3D ResNet} 
  
  \begin{tabular}{lllll}
  \toprule
    \textbf{Model Initialization} & \textbf{AUC} & \textbf{Precision}  & \textbf{Recall}  & \textbf{F1-score}\\
    \midrule
    Randomly & 0.706 ($\pm0.036$) & 0.344 ($\pm0.128$) & 0.407 ($\pm  0.058$) & 0.32 ($  \pm0.099$)\\ 
     Med3D \cite{chen2019med3d}  & 0.671 ($\pm0.044$) & \textbf{0.496} ($\pm0.069$) & 0.448 ($\pm0.062$) & 0.355 ($\pm0.087$)\\
     Instance-based CL \cite{ketabi2025multimodal}&  0.733 ($ \pm0.047$) &  0.491 ($\pm0.051$) &  0.502 ($\pm0.05$) & \textbf{0.48} ($\pm0.051$) \\ 
     Proposed MseaCL &  \textbf{0.743} ($ \pm0.036$) &  0.449($\pm0.148$) &  \textbf{0.536} ($\pm0.077$) & 0.454 ($\pm0.09$) \\ 
    \bottomrule
  \end{tabular}
  \label{table2} 
\end{table*}

\begin{table*}

  \centering 
  \caption{Mean External classification performance results of 3D ResNet } 
  \begin{tabular}{lllll}
  \toprule
    \textbf{Model Initialization} & \textbf{AUC} & \textbf{Precision}  & \textbf{Recall}  & \textbf{F1-score}\\
    \midrule
    Randomly & 0.479 ($\pm0.084$) & 0.191 ($\pm0.098$) & 0.352 ($\pm0.037$) & 0.229 ($\pm0.06$) \\ 
     Med3D \cite{chen2019med3d} & 0.405 ($\pm0.032$) & 0.158 ($\pm0.046$) & 0.32 ($\pm0.02$) & 0.202 ($\pm 0.023$)\\ 
     Instance-based CL \cite{ketabi2025multimodal}& 0.463 ($\pm0.057$) & 0.154 ($\pm0.083$) & 0.331 ($\pm0.005$) & 0.178 ($\pm0.054$)\\ 
     Proposed MseaCL & \textbf{0.689} ($\pm0.027$) & \textbf{0.406} ($\pm0.085$) & \textbf{0.441} ($\pm0.056$) & \textbf{0.358} ($\pm0.084$)\\ 
    \bottomrule
  \end{tabular}
  \label{table3} 
\end{table*}
\subsection{Explainability Results}



Tables \ref{table4} and \ref{table6} present the quantitative explainability assessment of the 3D ResNet model under different initialization techniques using 2D and 3D Dice scores between model attention maps and ground-truth segmentation masks. Across both internal and external settings, the proposed MSeaCL consistently achieved the highest explainability performance, outperforming all baselines by substantial margins.

In the internal evaluation, MSeaCL achieves a 2D Dice score of 28.8\%, which is notably higher than all baselines, nearly double that of instance-based CL (16.5\%) and more than ten times higher than randomly initialized and Med3D-based initialized models (1–3\%). Similarly, MSeaCL reaches a 3D dice score of 15.3\%, a significant improvement over the next-best method, instance-based CL (7.3\%). These results indicate that even in in-distribution settings, MSeaCL pretraining enables the model to more accurately localize tumor-relevant regions during classification.

Similar trends can be observed in the external evaluation, where models should generalize to images acquired from a different time period and scanner configuration. Baseline methods degrade substantially under this domain shift, with 3D Dice scores dropping to around 1\%, indicating almost no meaningful alignment between attention maps and segmentation masks. In contrast, MSeaCL achieves a 16.61\% 3D Dice score, representing a considerable improvement over the strongest baseline. The 2D Dice score shows similar subsantial enhancements, with MSeaCL reaching 27.94\%, surpassing the 1–5\% range achieved by the baselines. Importantly, these localization results were obtained without any segmentation supervision during training. Instead, attention maps were generated solely from a classification model and subsequently compared against expert-annotated segmentation masks. These findings highlight the important role of semantic-aware contrastive pretraining in producing attention maps that remain consistent and clinically relevant even under distribution shifts.

\begin{figure*}[ht!]
    \centering
    \begin{subfigure}{0.25\textwidth}
        \centering
        \includegraphics[width=0.6\textwidth]{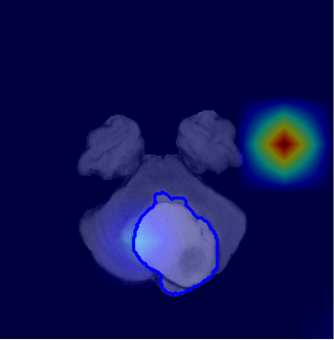}
        \caption{}
        \label{sc}
    \end{subfigure}
    \hfill
    \begin{subfigure}{0.25\textwidth}
        \centering
        \includegraphics[width=0.6\textwidth]{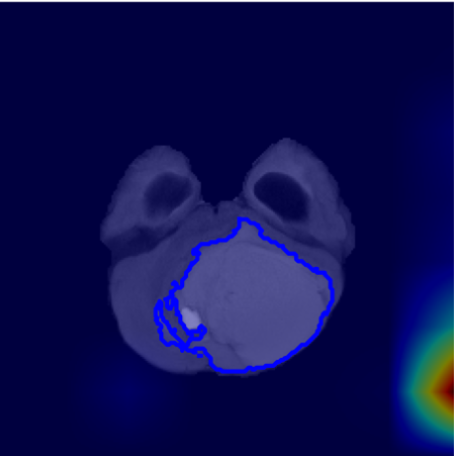}
        \caption{}
        \label{mednet}
    \end{subfigure}
    \hfill
    \begin{subfigure}{0.25\textwidth}
        \centering
        \includegraphics[width=0.6\textwidth]{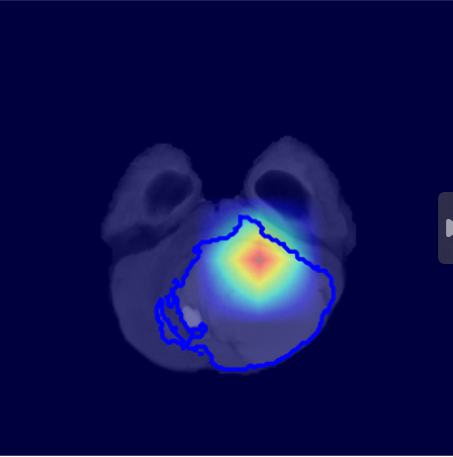}
        \caption{}
        \label{base_cl}
    \end{subfigure}
    \hfill
    \begin{subfigure}{0.25\textwidth}
        \centering
        \includegraphics[width=0.6\textwidth]{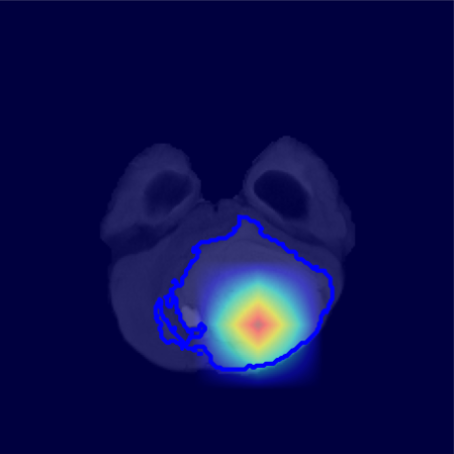}
        \caption{}
        \label{clsa}
    \end{subfigure}
    \caption{MRI with overlaid attention map and tumor contours for models initialized with: a: random weights b: Med3D \cite{chen2019med3d} c: instance-based CL \cite{ketabi2025multimodal} d:  MseaCL
}
    \label{result}
\end{figure*}

Figure \ref{result} provide qualitative comparison using a representative MRI slice from a sample in the external dataset. As shown, the model initialized with MSeaCL generates attention maps that largely overlap with the ground-truth tumor mask. In contrast, the baseline models tend to produce diffuse activations, often highlighting irrelevant regions or background areas. This illustrates that MSeaCL not only improves classification performance but also leads to more clinically meaningful explanations.

\begin{table}

  \centering 
  \caption{Mean Internal Test Classification Explainability  Results of 3D ResNet } 
  \begin{tabular}{lll}
  \toprule
    \textbf{Model Initialization} & \textbf{2D Dice Score} & \textbf{3D Dice Score}\\
    \midrule
    Randomly & 1.48\% 
 ($\pm2.52\%$)  & 0.6\% ($\pm1.03\%$)\\ 
    Med3D \cite{chen2019med3d}& 2.9\% ($\pm4.3\%$)& 1.4\% ($\pm2.3\%$) \\ 
    Instance-based CL \cite{ketabi2025multimodal}& 16.5\% ($\pm12.4\%$) & 7.3\% ($\pm5.5\%$)\\ 
    Proposed MseaCL & \textbf{28.8\%} ($\pm6\%$) & \textbf{15.3\%} ($\pm3.7\%$)\\ 
    \bottomrule
  \end{tabular}
  \label{table4} 
\end{table}
\begin{table}

  \centering 
  \caption{Mean External Classification Explainability results of 3D ResNet} 
  \begin{tabular}{lll}
  \toprule
    \textbf{ Initialization} &  \textbf{2D Dice Score} & \textbf{3D Dice Score}\\
    \midrule
    Randomly & 2.74\% ($\pm3.99\%$)  &  1.33\% ($\pm2.12\%$)\\ 
     Med3D \cite{chen2019med3d} &  1.41\% ($\pm1.97\%$) & 0.63\% ($\pm0.88\%$)\\ 
    typical CL \cite{ketabi2025multimodal}& 4.25\% ($\pm4.33\%$) & 1.57\% ($\pm1.7\%$)\\ 
    Proposed MseaCL & \textbf{27.94}\% ($\pm0.66\%$) & \textbf{16.61}\% ($\pm0.35\%$)\\ 
    \bottomrule
  \end{tabular}
  \label{table6} 
\end{table}

\subsection{Ablation Studies}


To better understand the contribution of each component of MseaCL, we conducted a number of ablation studies and evaluated their impact on the external performance using both AUC (primary metric) and F1-score (secondary metric). Specifically, we examined two key modifications. First, we replaced the margin-based contrastive loss used in MseaCL with a standard InfoNCE-style loss to evaluate the importance of explicit margin separation. Second, we increased the base margin in our original loss function from 0.5 to 0.75 to assess the sensitivity of the method to margin scaling.

As summarized in Table \ref{ablation}, both experiments led to a decrease in AUC relative to the full MseaCL framework, demonstrating that the original margin formulation and its corresponding margin value are essential for maximizing discriminative performance. Interestingly, while the InfoNCE experiment yielded an improvement in F1-score, this came at the cost of a notable reduction in AUC, indicating that the model became less reliable at ranking positive and negative samples across the full decision curve. Overall, these findings highlight that MseaCL’s margin-based contrastive design plays a critical role in learning robust representations.

\begin{table}[ht!]

  \centering 
  \caption{Mean External Classification Results of the Ablation Experiments}
  \begin{tabular}{lll}
  
  \toprule
    \textbf{Experiment} & \textbf{Mean AUC} & 
    \textbf{Mean F1-score}\\
    \midrule
    Base Margin Change  & 0.647 & 0.296\\ 
    InfoNCE Loss & 0.567 & \textbf{0.434} \\ 
    Proposed MseaCL & \textbf{0.661} & 0.343 \\
   
    \bottomrule
  \end{tabular}
  \label{ablation} 
\end{table}

%% file: sec/5_discussion.tex
\section{Discussion}

In this chapter, we introduced a semantic-aware CL framework, MSeaCL, designed to address the issue of false negatives in multimodal representation learning. By integrating 3D MRI scans with corresponding radiology reports, the proposed framework leverages semantic information from clinical text to guide the alignment of image–text representations. Specifically, MSeaCL computes similarity scores between radiology reports and uses these scores to adaptively adjust the contribution of negative image–report pairs in the contrastive objective. This mechanism reduces the penalization of semantically similar samples, which would otherwise be incorrectly treated as dissimilar under conventional instance-based CL.

Our results demonstrate that this semantic-aware formulation leads to consistent improvements in downstream performance. When fine-tuned on the task of three-class pLGG genetic marker classification, MSeaCL outperforms all baseline methods based on AUC, including standard CL and other initialization strategies. Notably, the performance gains are particularly significant in the external validation setting, where the proposed model achieves substantial improvements over baseline methods. This observation suggests that mitigating false negatives during representation learning enhances the model’s ability to capture robust, clinically meaningful features that generalize across variations in data distributions, such as differences in scanners, acquisition protocols, and time frames.

Beyond classification performance, the benefits of MSeaCL are also evident from the explainability analysis. Incorporating MSeaCL-pretrained weights into the 3D ResNet model significantly improves the alignment between model-generated attention maps and tumor-related regions. This improvement indicates that the learned representations are more closely aligned with clinically relevant structures, rather than being influenced by spurious or background noises. In contrast to standard CL, which enforces strict separation between all non-matching pairs, the semantic-aware loss encourages the model to preserve meaningful similarities across samples that share clinical attributes. As a result, the model is better able to focus on anatomically and pathologically relevant regions when making predictions.


The observed improvements in both performance and explainability suggest that semantic alignment plays a dual role in MML. On one hand, it enhances discriminative capability by refining the embedding space; on the other hand, it promotes explainability by encouraging the model to rely on clinically meaningful features. This dual benefit is especially valuable in medical settings, where both performance and explainability are essential for deployment in real-world clinical workflows.

Despite these promising results, several limitations remain. First, the current framework relies on report-level semantic similarity, which may not fully capture detailed clinical attributes present in localized image regions. Incorporating other forms of guidance, such as concept-level alignment, could further improve both representation quality and explainability. Second, the choice of similarity metric and weighting strategy introduces additional hyperparameters that may influence model performance and require careful tuning. Finally, while improved attention alignment provides evidence of enhanced explainability, further validation through human-centered evaluation and user studies would be necessary to assess the practical utility of the generated explanations.

%% file: sec/6_conclusion.tex
\section{Conclusion and Future Work}
In this study, we proposed a semantic-aware MRI–report CL framework that reduces false negatives by adjusting distances between mismatched image–report pairs based on report-level semantic similarity. Our results show that this approach improves the generalizability of instance-based CL and other baselines by at least 22.6\% in multiclass pediatric tumor genetic marker classification. Applying this framework provides more clinically meaningful representations with higher generalizability across distribution shifts. As future work, we can enhance representation quality by exploring finer-grained semantic references beyond full reports, such as high-level clinical concepts.